\documentclass[letterpaper, 10 pt, conference]{ieeeconf}  
\IEEEoverridecommandlockouts                              
\usepackage{graphicx}
\usepackage{bm} 
\usepackage{makecell} 

\usepackage{epsfig} 
\usepackage{amsmath} 
\usepackage{amssymb}  
\usepackage{algorithm}
\usepackage{algpseudocode}
\usepackage{multirow}
\usepackage{cite}
\makeatletter
\let\NAT@parse\undefined
\makeatother
\usepackage[
            citecolor=black,
            ]{hyperref}
\usepackage{booktabs}
\usepackage{tabularx} 
\usepackage{lineno,hyperref}
\usepackage{color,xcolor}
\usepackage{tabularx}
\usepackage{amsmath}
\usepackage[skip=3pt, font={small}]{caption}

\renewcommand{\thetable}{\Roman{table}}  
\title{\LARGE \bf
NavCrafter: Exploring 3D Scenes from a Single Image
}

\author{
    Hongbo Duan$^{1,2}$, Peiyu Zhuang$^{3}$, Yi Liu$^{1}$,  Zhengyang Zhang$^{1}$, Yuxin Zhang$^{1}$, Pengting Luo$^{4}$, Fangming Liu$^{2}$, \\
    Xueqian Wang$^{1}$
    \thanks{
    This work was supported by the National Natural Science Foundation of China under Grant Nos. 62293545 and U21B6002, in part by the Major Key Project of PCL under Grant PCL2024A06 and PCL2025A10, and in part by the Shenzhen Science and Technology Program under Grant RCJC20231211085918010.
    \textit{(Corresponding author: Xueqian Wang.)}}
    \thanks{$^{1}$ Center for Artificial Intelligence and Robotics, Shenzhen International Graduate School, Tsinghua University, Shenzhen 518055, China (e-mail: 
    \href{mailto:dhb24@mails.tsinghua.edu.cn}{dhb24@mails.tsinghua.edu.cn};
    \href{mailto:wang.xq@sz.tsinghua.edu.cn}{wang.xq@sz.tsinghua.edu.cn})}
    \thanks{$^{2}$ Peng Cheng Laboratory, 518108, China}
    \thanks{$^{3}$ School of Cyber Science and Technology, Shenzhen Campus of Sun Yat-sen University, China}
    \thanks{$^{4}$ Central Media Technology Institute, Huawei
Incorporated Company, China.}
}

\makeatletter
\let\@oldmaketitle\@maketitle
\renewcommand{\@maketitle}{\@oldmaketitle
	\vspace{0.2cm}
	\centering
	\setcounter{figure}{0}
	\begin{minipage}{1.0\linewidth}
		\includegraphics[width=1.0\textwidth]{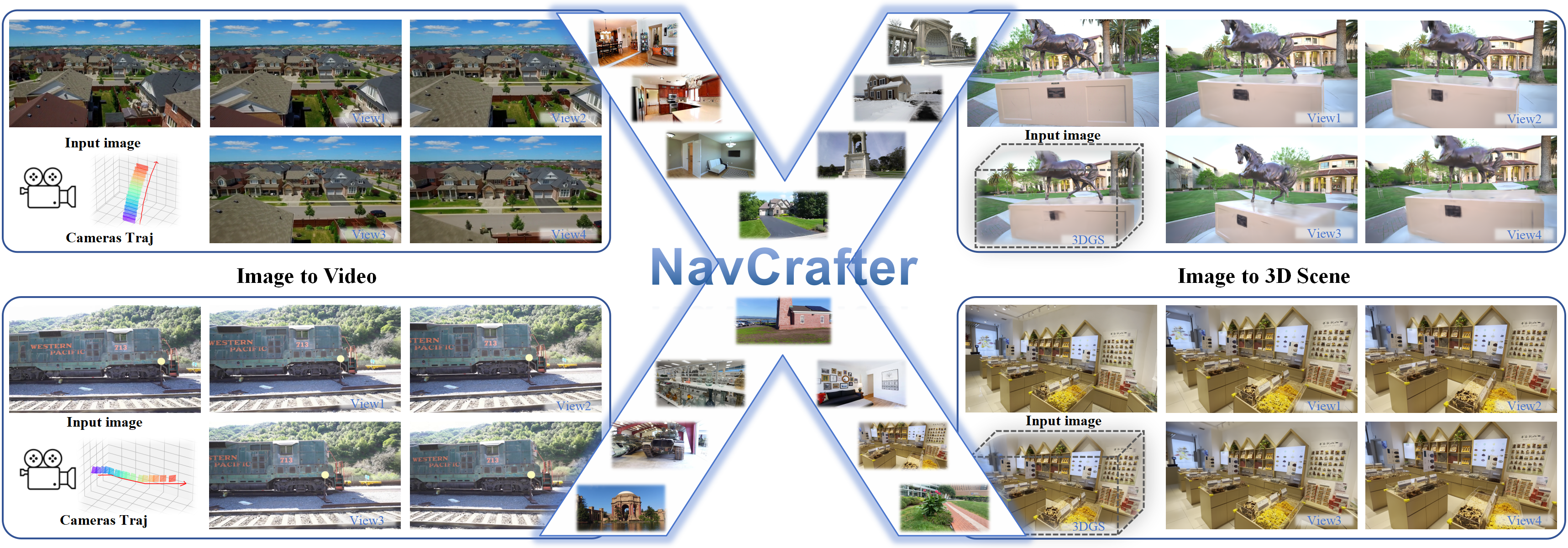}
		\vspace{0cm}
		\captionof{figure}{\label{fig:1} Visual results generated by NavCrafter. Given a single image, NavCrafter reconstructs 3D scenes from the camera-guided video diffusion model.}
	\end{minipage}
	\vspace{-0.3cm}
}
\makeatother

\begin{document}

\maketitle
\thispagestyle{empty}
\pagestyle{empty}

\begin{abstract}
Creating flexible 3D scenes from a single image is vital when direct 3D data acquisition is costly or impractical. 
We introduce \textit{NavCrafter}, a novel framework that explores 3D scenes from a single image by synthesizing novel-view video sequences with camera controllability and temporal-spatial consistency. 
NavCrafter leverages video diffusion models to capture rich 3D priors and adopts a geometry-aware expansion strategy to progressively extend scene coverage. 
To enable controllable multi-view synthesis, we introduce a \textit{multi-stage camera control mechanism} that conditions diffusion models with diverse trajectories via dual-branch camera injection and attention modulation.
We further propose a \textit{collision-aware camera trajectory planner} and an \textit{enhanced 3D Gaussian Splatting (3DGS) pipeline} with  depth-aligned supervision, structural regularization and refinement. 
Extensive experiments demonstrate that NavCrafter achieves state-of-the-art novel-view synthesis under large viewpoint shifts and substantially improves 3D reconstruction fidelity.
\end{abstract}

\section{Introduction}

Humans naturally perceive 3D structures from a single image, effortlessly estimating depth, inferring spatial layouts, and reasoning about occluded regions. Emulating this ability in computational models---i.e., generating flexible 3D scenes from sparse or even a single observation---has transformative potential for domains where direct 3D capture is expensive or infeasible, including filmmaking, VR/AR, robotics, and social platforms.

Recent advances in learnable scene representations, such as Neural Radiance Fields (NeRF)~\cite{mildenhall2021nerf} and 3DGS~\cite{kerbl20233d}, have enabled photorealistic rendering of 3D scenes. However, these methods typically require dense multi-view inputs, severely limiting their applicability in scenarios with restricted observations. A more practical yet challenging setting involves synthesizing novel views (NVS) from a single image, which requires comprehensive understanding of 3D structure, appearance, semantics, and occlusion reasoning.

Generative models, especially diffusion models~\cite{ho2020denoising,song2020score}, provide principled solutions for novel-view synthesis (NVS). 
Image-based methods~\cite{chung2023luciddreamer,yu2025wonderworld} often accumulate geometric errors, 
while video-based models~\cite{blattmann2023stable,wang2024motionctrl} struggle with dynamic content and weak camera supervision. 
Recent video generation approaches~\cite{he2025cameractrl,ren2025gen3c} achieve impressive realism by learning distributions of real-world videos, 
yet their application to NVS remains limited by two persistent challenges: (1) {controllability}—explicit specification of camera motions and scene composition; 
and (2) {consistency}—maintaining spatio-temporal coherence across long sequences for reliable 3D reconstruction. 
Although some works attempt to address these issues through fine-tuning with additional images, text prompts, or camera parameters~\cite{wang2024motionctrl,chen2024liftimage3d}, 
precise control of complex trajectories and consistent synthesis under large and rapid viewpoint changes remain unsolved, often leading to geometric inconsistencies that degrade both 3DGS reconstruction quality and Structure-from-Motion (SfM) pose estimation.

To overcome these limitations, we propose \textit{NavCrafter}, a framework for {exploring 3D scenes from a single image} (Fig. ~\ref{fig:1}). 
NavCrafter leverages rich 3D priors from video diffusion models and employs a geometry-aware expansion process to progressively integrate novel content into a global scene structure. 
This design enables precise camera control, broad scene coverage, and high-fidelity 3D reconstruction.

Our contributions are summarized as follows:
\begin{itemize}
    \item \textbf{Controllable Novel View Synthesis:} We propose a multi-stage camera control architecture that incorporates camera trajectories into video diffusion models via dual-branch camera injection and attention modulation.
    \item \textbf{Iterative View Synthesis with Collision-Aware Camera Trajectory Planning:} We present an iterative NVS strategy with collision-aware camera trajectory planning, progressively extending the coverage of synthesized views and reconstructed point clouds.
    \item \textbf{Geometry-aware 3D Reconstruction:} We enhance the 3DGS reconstruction pipeline with depth-aligned supervision, structural regularization and refinement to improve geometric consistency.
\end{itemize}

Extensive experiments show that NavCrafter achieves high-quality novel view synthesis under challenging viewpoint changes and significantly improves downstream 3D scene reconstruction.

\section{Related Work}

\subsection{Novel View Synthesis (NVS)}
Generating novel views from a set of posed images has been extensively studied~\cite{mildenhall2021nerf,kerbl20233d}. However, most methods require dense input views and often produce severe artifacts when extrapolating to extreme viewpoints.  To mitigate these limitations, 
several approaches introduced geometric priors for regularization~\cite{park2025dropgaussian,li2024dngaussian}, 
but their performance is sensitive to noise in depth or normal estimates. 
Feedforward models have been explored to directly predict novel views from sparse inputs~\cite{chen2024mvsplat360}, 
yet they are constrained by the scarcity of training data and struggle to generalize to unseen domains and large viewpoint shifts.  

With the rise of image and video generation models, See3D~\cite{ma2025you} and CAT3D~\cite{gao2024cat3d} introduced generative priors to improve sparse-view NVS, 
though their per-scene optimization remains computationally expensive. 
More recent approaches utilize video diffusion models and global point clouds to improve multi-view consistency~\cite{chen2024liftimage3d,yu2024viewcrafter}, 
but their effectiveness depends on point cloud quality and remains limited to narrow-scoped scenes. 

\subsection{Camera-Conditioned Video Diffusion Models}
Camera-conditioned video diffusion models have recently attracted growing attention~\cite{wang2024motionctrl,xiao20243dtrajmaster,he2025cameractrl}. 
Early works explored training-free conditioning strategies~\cite{hu2024motionmaster} or integrated LoRA modules~\cite{hu2022lora} into diffusion pipelines for limited forms of camera control. 
Recent efforts, such as Gen3C~\cite{ren2025gen3c}, incorporated ControlNet-like conditioning with cross-attention mechanisms, but due to high computational costs, pose control was only applied at low-resolution stages in cascaded generators. 
Methods like DimensionX~\cite{sun2024dimensionx} achieved basic control via multiple LoRA modules but struggled with complex motions. 
Wonderland~\cite{liang2025wonderland} and StarGen~\cite{zhai2025stargen} synthesize videos from a single view and trajectory but cannot supplement existing 3D structures, 
limiting scene coverage. Similarly, See3D~\cite{ma2025you} and ViewCrafter~\cite{yu2024viewcrafter} can inpaint missing perspectives but fail to handle large viewpoint changes.  
In contrast, our method introduces a multi-stage camera control mechanism directly into the video diffusion backbone, enabling precise pose control while preserving generation quality.

\subsection{3D Scene Generation}
While object-level 3D generation~\cite{gao2024cat3d} has made remarkable progress, full-scene generation remains underexplored. 
Early works~\cite{chung2023luciddreamer, yu2024wonderjourney} combined monocular depth warping with diffusion-based inpainting, 
but depth estimation errors and per-view refinements often led to distortions and inconsistent geometry. 
Others explored video diffusion models coupled with point clouds~\cite{yu2024viewcrafter,ren2025gen3c}, 
which improved multi-view consistency but remained constrained to narrow-range scenes due to reliance on point cloud quality. 
Our work departs from these approaches by explicitly embedding camera control into the video diffusion backbone and coupling it with a collision-aware camera trajectory planning strategy. 
This enables progressive scene expansion and, together with an enhanced 3DGS-based reconstruction pipeline, allows us to generate wide-scope, high-fidelity 3D scenes from a single image.

\begin{figure*}[t] 
    \centering
    \includegraphics[width=1.0\textwidth]{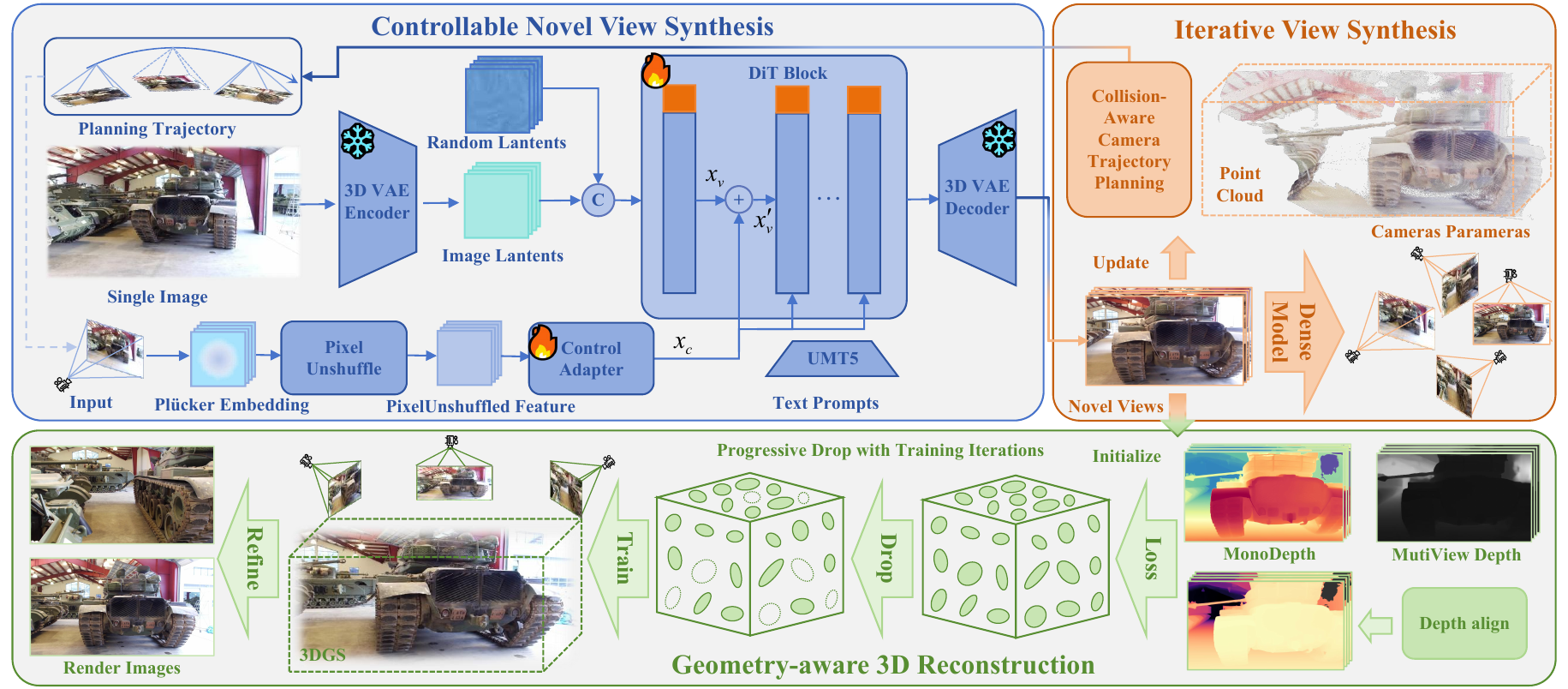} 
    \caption{The NavCrafter framework consists of three modules: 
    (1) Controllable novel-view synthesis via video diffusion, integrating camera trajectories to control video generation and achieve temporally consistent novel views;
    (2) Iterative view synthesis with collision-aware camera trajectory planning, avoiding scene collisions and optimizing camera trajectories;
    (3) Geometry-aware 3D reconstruction with enhanced 3D Gaussian Splatting, incorporating depth-aligned supervision, structural regularization and image diffusion model refinement.}    
    \label{fig:framework}
\end{figure*}

\section{Preliminaries} 

\subsection{Video Diffusion Model} 
A diffusion model \cite{ho2020denoising,song2020score} consists of a forward and a denoising process. 
In the forward process, the diffusion model gradually adds Gaussian noise to a clean image $x_0$ from time $0$ to $T$. 
The noisy image $x_t$ at a certain time $t \in [0, T]$ can be expressed as $x_t = \alpha_t x_0 + \sigma_t \epsilon$, where $\alpha_t$ and $\sigma_t$ are predefined hyperparameters. 
In the denoising process, a noise predictor $\epsilon_\theta(x_t, t)$ with parameters $\theta$ is trained to predict noise in $x_t$ for generation. 
Given the corresponding condition $y$ for $x$, the training objective of a diffusion model is:
\begin{equation}
\min_\theta \mathbb{E}_{t \sim \mathcal{U}(0,1), \epsilon \sim \mathcal{N}(0,I)} \left[ \| \epsilon_\theta(x_t, t; y) - \epsilon \|_2^2 \right].
\label{eq:diffusion_objective}
\end{equation}

Recent video diffusion models \cite{wan2025wan} typically employ a 3D-VAE encoder $\bm{\mathit{E}}$ to compress the source video into a latent space where the diffusion model is trained. 
The generated latent video is subsequently decoded to the pixel space using the corresponding decoder $\bm{\mathit{D}}$.

\subsection{3D Gaussian Splatting} 

3DGS represents a scene with a set of 3D Gaussians, each defined by a center $\boldsymbol{\mu} \in \mathbb{R}^3$, color $\mathbf{c} \in \mathbb{R}^3$, opacity $\eta$, scale $\mathbf{S} \in \mathbb{R}^{3 \times 3}$, and rotation $\mathbf{R} \in \text{SO}(3)$. 
The rendering color along a ray $\mathbf{r}$ is computed by standard volume rendering:
\begin{equation}
\mathcal{C}(\mathbf{p}) = \sum_{i=1}^N \alpha_i \mathbf{c}_i \prod_{j=1}^{i-1} (1 - \alpha_j),
\end{equation}
where $\alpha_i$ is the opacity of Gaussian $i$. 
For a point $\mathbf{p}$, $\alpha_i$ is given by
\begin{equation}
\alpha_i = \eta \exp\!\left( -\tfrac{1}{2} (\mathbf{p}-\boldsymbol{\mu})^\top 
\Sigma^{-1} (\mathbf{p}-\boldsymbol{\mu}) \right),
\end{equation}
with covariance $\Sigma = \mathbf{R}\mathbf{S}\mathbf{S}^\top\mathbf{R}^\top$.

\section{Methodology}

\subsection{System Overview}

As shown in Fig. ~\ref{fig:framework}, {NavCrafter} consists of three modules:  

\paragraph{Module I: Controllable Novel View Synthesis}  
In Sec. \ref{NVS}, from a single input image, we employ a video diffusion model with multi-stage camera control, 
which incorporates camera trajectories to ensure precise viewpoint control and temporal consistency across synthesized views.

\paragraph{Module II: Iterative View Synthesis with Collision-Aware Camera Trajectory Planning}  
In Sec. \ref{vggt}, we propose a collision-aware trajectory planning module that iteratively explores novel views, avoiding scene collisions and correcting camera trajectories.

\paragraph{Module III: Geometry-aware 3D Reconstruction}  
In Sec. \ref{3dgs}, synthesized views with poses are reconstructed via enhanced 3D Gaussian Splatting, further refined by:  
(1) depth-aligned supervision for geometric accuracy, 
(2) structural regularization to mitigate overfitting, and
(3) image diffusion-based refinement for visual fidelity.

\subsection{Controllable Novel View Synthesis}
\label{NVS}
Video diffusion models lack explicit camera trajectory control, limiting 3D reconstruction for static scenes. We propose a framework integrating precise pose information to enable multi-view-consistent synthesis with 3D-aware latents.

\subsubsection{Camera Trajectory Representation}
Per-pixel rays are derived from frame $f$'s camera parameters ($\mathbf{R}_f \in \mathbb{R}^{3\times 3}$, $\mathbf{t}_f \in \mathbb{R}^3$, $\mathbf{K}_f \in \mathbb{R}^{3\times 3}$). The normalized ray direction at $(u_f,v_f)$ is:
\begin{equation}
\mathbf{d}_{u_f,v_f} = \frac{\mathbf{R}_f \mathbf{K}_f^{-1} [u_f, v_f, 1]^T + \mathbf{t}_f}{\|\mathbf{R}_f \mathbf{K}_f^{-1} [u_f, v_f, 1]^T + \mathbf{t}_f\|}.
\end{equation}

Plücker embedding encodes ray orientation and camera center:
\begin{equation}
\dot{\mathbf{p}}_{u_f,v_f} = (\mathbf{t}_f \times \mathbf{d}_{u_f,v_f}, \mathbf{d}_{u_f,v_f}) \in \mathbb{R}^6.
\end{equation}
Stacking over frames yields $\mathbf{p} \in \mathbb{R}^{T \times H \times W \times 6}$, precomputed offline for efficiency.

\subsubsection{Multi-Stage Camera Control Architecture}
We propose a three-stage architecture to achieve persistent camera guidance without full fine-tuning.

\textbf{Dual-Branch Camera Injection:}  
Downsampled trajectory embeddings $\dot{\mathbf{p}}_{u,v}$ are encoded by a 3D convolutional adapter $\mathcal{A}$ into ${x}_{c}$:
\begin{equation}
{x}_{c} = \mathcal{A}(\dot{\mathbf{p}}_{u,v}).
\end{equation}
${x}_{c}$ is injected into the video tokens ${x}_v$ before each Diffusion Transformer (DiT) block as ${x}_v' = {x}_v + {x}_{c}$ to initialize trajectory constraints, and is also added to self-attention outputs as ${x}_{a}' = {x}_{a} + {x}_{c}$ to reinforce signals. Random reference frames in self-attention further enhance cross-view consistency.

\textbf{LoRA Attention Modulation:}  
Trajectory embeddings are also projected by a lightweight 3D convolutional encoder into LoRA control tokens ${x}_{l}$, which share the same dimension as video tokens. These tokens modulate the query, key, and value projections, e.g.,
\begin{equation}
{Q}' = {Q} + \alpha \cdot {W}_{u}({W}_{d} \cdot {x}_{l}),
\end{equation}
where ${W}_{u}$ and ${W}_{d}$ are low-rank matrices and $\alpha$ controls the modulation strength. This directs attention toward trajectory-consistent regions.  

Overall, the integration of dual-branch feature injection and attention-level modulation ensures accurate trajectory following, enhanced geometric consistency, and efficient adaptation without full retraining. Training details are given in Sec.~\ref{section IV}.

\subsection{Iterative View Synthesis with Collision-Aware Camera Trajectory Planning}
\label{vggt}

To mitigate instability and high cost in long-horizon video diffusion, we adopt iterative view synthesis with collision-aware camera trajectory planning. 

Given a sequence of RGB images \((I_i)_{i=1}^N\), where each \(I_i \in \mathbb{R}^{3 \times H \times W}\) observes the same 3D scene, 
VGGT~\cite{wang2025vggt} uses a transformer $\mathcal{D}(\cdot)$ to generate 3D annotations for each frame:

\begin{equation}
\mathcal{D}\left((I_i)_{i=1}^N\right) = (\mathbf{g}_i, \mathbf{d}_i, \mathbf{p}_i)_{i=1}^N,
\end{equation}
where \(\mathbf{g}_i \in \mathbb{R}^9\) represents camera intrinsics and extrinsics, 
\(\mathbf{d}_i \in \mathbb{R}^{H \times W}\) is the depth map, and \(\mathbf{p}_i \in \mathbb{R}^{3 \times H \times W}\) is the point cloud. 
The point cloud is defined in the coordinate system of the first camera \(\mathbf{g}_1\), which serves as the world reference frame.

Starting from the reference point cloud $\mathcal{P}_\text{ref}$, the camera iteratively moves from the current pose $\mathcal{C}_\text{curr}$ to selected next-best-views (NBVs).

\subsubsection{Collision-Aware NBV Selection}  
At each iteration, $K$ candidate poses are sampled around $\mathcal{C}_\text{curr}$.  
Colliding poses are removed by $\mathcal{G}(\cdot)$, and valid ones are scored with $\mathcal{F}(\cdot)$ using a visibility mask $\mathcal{M}$ from point-cloud rendering, favoring informative and less occluded views.

\subsubsection{Adaptive Trajectory Generation} 
The optimal pose $\mathcal{C}_\text{nbv}$ is selected via spherical interpolation. 
If the interpolated trajectory $\mathcal{T}_\text{smooth}$ intersects the scene, 
continuous collision-aware optimization adjusts the trajectory by minimizing a combined cost of collision risk and trajectory smoothness:

\begin{equation}
\begin{split}
\min_{\mathcal{T}_t} \; & \sum_t \max\big(0, r_\text{safe} - d(\mathcal{T}_t, \mathcal{P}_\text{curr})\big) \\
& + \lambda \sum_t \|\mathcal{T}_{t+1} - \mathcal{T}_t\|^2,
\end{split}
\end{equation}
where $r_\text{safe}$ is the safety radius, $\text{dist}(\mathcal{T}_t, \mathcal{P}_\text{curr})$ computes the shortest distance from the trajectory point $\mathcal{T}_t$ to the current point cloud $\mathcal{P}_\text{curr}$, and $\lambda$ controls trajectory smoothness. This formulation ensures collision-free, smooth, and physically plausible camera trajectories.

\subsubsection{Progressive Scene Enhancement} 
Synthesized views $\mathcal{I}_\text{nbv}$ by NavCrafter in Sec.~\ref{NVS} using $\mathcal{V}(\cdot)$ are back-projected via $\mathcal{D}(\cdot)$ to progressively expand coverage and refine reconstruction. This iterative process continues until $N$ poses are generated.

The procedure is summarized in Algorithm~\ref{alg:collision_aware_camera_planning}.

\begin{algorithm}
\caption{Collision-Aware Camera Trajectory Planning}
\label{alg:collision_aware_camera_planning}

\begin{algorithmic}[1]
\State \textbf{Initialize} scene center $\mathbf{o}$, current point cloud $\mathcal{P}_\text{curr} \leftarrow \mathcal{P}_\text{ref}$, 
current camera pose $\mathcal{C}_\text{curr} \leftarrow \mathcal{C}_\text{ref}$, 
collision detector $\mathcal{G} \leftarrow \text{initialize}(\mathcal{P}_\text{ref})$, $step \leftarrow 0$

\While{$step \leq N$}
    
    \State Spherically sample $K$ candidate poses $\mathcal{C}_\text{can} = \{\mathcal{C}_\text{can}^1, \ldots, \mathcal{C}_\text{can}^K\}$ from the searching space $\mathcal{S}$ around the current pose $\mathcal{C}_\text{curr}$, initialize candidate mask set $\mathcal{M}_\text{can} = \{\}$
    
    \For{$\mathcal{C}$ in $\{\mathcal{C}_\text{can}^1, \ldots, \mathcal{C}_\text{can}^K\}$}
        \If{not $\mathcal{G}(\mathcal{C})$}  
            \State $\mathcal{M}_\mathcal{C} = \textit{Render}(\mathcal{P}_\text{curr}, \mathcal{C})$
            \State$\mathcal{M}_\text{can}.\textit{append}(\mathcal{M}_\mathcal{C})$
        \Else
            \State $\mathcal{M}_\text{can}.\textit{append}(\emptyset)$ \Comment{Collision}
        \EndIf
    \EndFor
    
    \State $\mathcal{C}_\text{nbv} = \arg\max\limits_{\mathcal{C} \in \mathcal{C}_\text{can}} \mathcal{F}(\mathcal{C})$
    
    \State $\mathcal{T}_\text{smooth} = \textit{SphericalInterpolate}(\mathcal{C}_\text{curr}, \mathcal{C}_\text{nbv})$
    
    \If{$\mathcal{G}(\mathcal{T}_\text{smooth})$}
        \State $\mathcal{T}_\text{smooth} = \textit{CollisionOptimization}(\mathcal{T}_\text{smooth}, \mathbf{o})$
    \EndIf
    
    \State $\mathcal{I}_\text{nbv} = \mathcal{V}(\mathcal{T}_\text{smooth}, \mathcal{P}_\text{curr})$ ,  $\mathcal{P}_\text{curr} \leftarrow \mathcal{D}(\mathcal{I}_\text{nbv}, \mathcal{P}_\text{curr})$
    
    \State $\mathcal{C}_\text{curr} \leftarrow \mathcal{C}_\text{nbv}$, $step \leftarrow step + 1$
\EndWhile

\State \Return 
\end{algorithmic}
\end{algorithm}

\subsection{Geometry-aware 3D Reconstruction}
\label{3dgs}
\begin{figure*}[t] 
    \centering
    \includegraphics[width=1.0\textwidth]{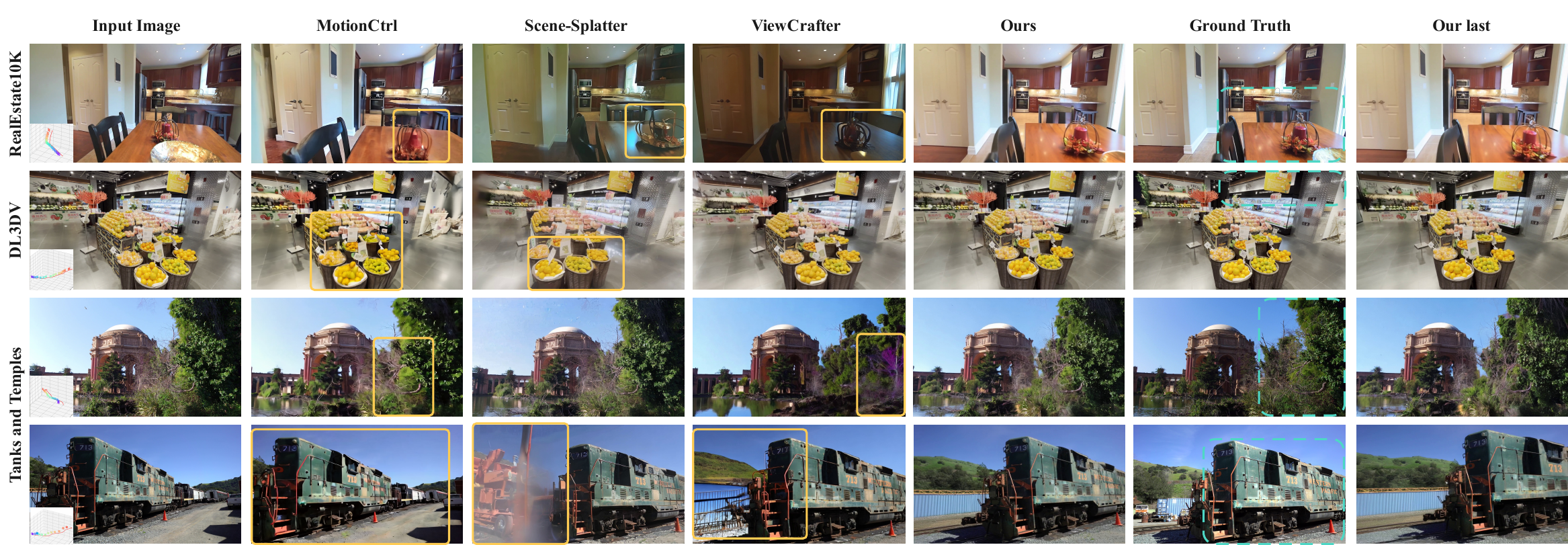} 
    \caption{Qualitative comparison with prior methods in controllable novel view synthesis, where the first column shows the input image and camera trajectory. Blue bounding boxes indicate reference areas for easier comparison, while orange ones highlight low-quality generations.}
    \label{fig:qualitative}
\end{figure*}
In this section, we focus on geometry-aware 3D scene reconstruction, 
composed of three key components: depth-aligned supervision, structural regularization, 
and multi-view refinement for high-fidelity, geometrically consistent reconstructions.
These camera parameters and multi-view point clouds which obtained in Sec. \ref{vggt} are used for 3DGS initialization.

\subsubsection{Depth-aligned Supervision}  
We use depth supervision to improve geometric consistency. To improve depth quality, we combine absolute depths from neural matching with monocular depth predictions. 
The depth estimated by VGGT~\cite{wang2025vggt} is not accurate enough but is aligned with camera poses, while monocular predictions perform better on edges. 
We calibrate relative monocular depths $\mathbf{d}_m$ obtained from MoGe-2~\cite{wang2025moge} against absolute depths $\mathbf{d}_v$ by solving:
\begin{equation}
\begin{aligned}
\min_{\textit{scale},\textit{bias}} 
\Big\| \mathcal{M} \cdot \Big( \hat{\mathbf{d}}_m - \frac{1}{\mathbf{d}_v} \Big) \Big\|^2, 
\hat{\mathbf{d}}_m = \frac{\textit{scale}}{\mathbf{d}_m} + \textit{bias},
\end{aligned}
\end{equation}
where $\textit{scale}$ and $\textit{bias}$ are the calibration parameters, 
$\hat{\mathbf{d}}_m$ denotes the calibrated monocular depth, and $\mathcal{M}$ is a mask for valid non-sky regions.

\subsubsection{Structural Regularization}  
To mitigate overfitting under sparse views, we use DropGaussian\cite{park2025dropgaussian}, which randomly removes Gaussians with dropping rate \(r\). The opacity value of the remaining Gaussians as follows:
\begin{equation}
\tilde{o}_i = M(i) \cdot o_i, \quad
M(i) = \frac{1}{1-r} \cdot \mathbb{I}_{\text{keep}}(i),
\end{equation}
where \(\mathbb{I}_{\text{keep}}(i)\) indicates whether Gaussian \(i\) is kept. We apply a progressive dropping schedule:
\begin{equation}
r_t = \gamma \cdot \frac{t}{t_{\text{total}}},
\end{equation}
where \(t\) is the current iteration, \(t_{\text{total}}\) is the total number of iterations, and \(\gamma\) is the maximum dropping rate.

\subsubsection{Loss Function Design}  
We use a multi-constraint loss function to balance photometric fidelity and geometric consistency. 
L1 RGB loss \(\mathcal{L}_{\text{1RGB}}\) and perceptual loss \(\mathcal{L}_{\text{lpips}}\) ensure texture accuracy, 
while L1 depth loss \(\mathcal{L}_{\text{1depth}}\), based on the calibrated depth $\hat{\mathbf{d}}_m$, ensures 3D consistency:
\begin{equation}
\mathcal{L} = \mathcal{L}_{\text{1RGB}} + \mathcal{L}_{\text{lpips}} + \mathcal{L}_{\text{1depth}}.
\label{eq:total_loss}
\end{equation}

\subsubsection{Refinement}  
For improved visual fidelity, multi-view images \(I\) are rendered and perturbed with noise, then iteratively denoised using the image diffusion model Difix3D+~\cite{wu2025difix3d+}:
\begin{equation}
\hat{I} = f_\theta(\alpha_t I + \sigma_t \epsilon, t),
\end{equation}
where \(t\) denotes the diffusion timestep and \(f_\theta\) denotes the image diffusion model. 
The resulting images \(\hat{I}\) serve as additional supervision for 3DGS reconstruction and are optimized with the same loss in Eq.~\eqref{eq:total_loss}.

\section{Experiments and Results}
In this section, we first describe implementation details in Section~\ref{section:implementation}. 
Quantitative and qualitative results for controllable video generation and 3D scene reconstruction are presented in Sections~\ref{subsectionB} and~\ref{subsection:3d_reconstruction}, respectively. 
Collision-aware camera planning and ablation studies are analyzed in Sections~\ref{subsection:collision_planning} and~\ref{subsection:ablation}.

\label{section IV}
\begin{figure*}[ht] 
    \centering
    \includegraphics[width=0.8\textwidth]{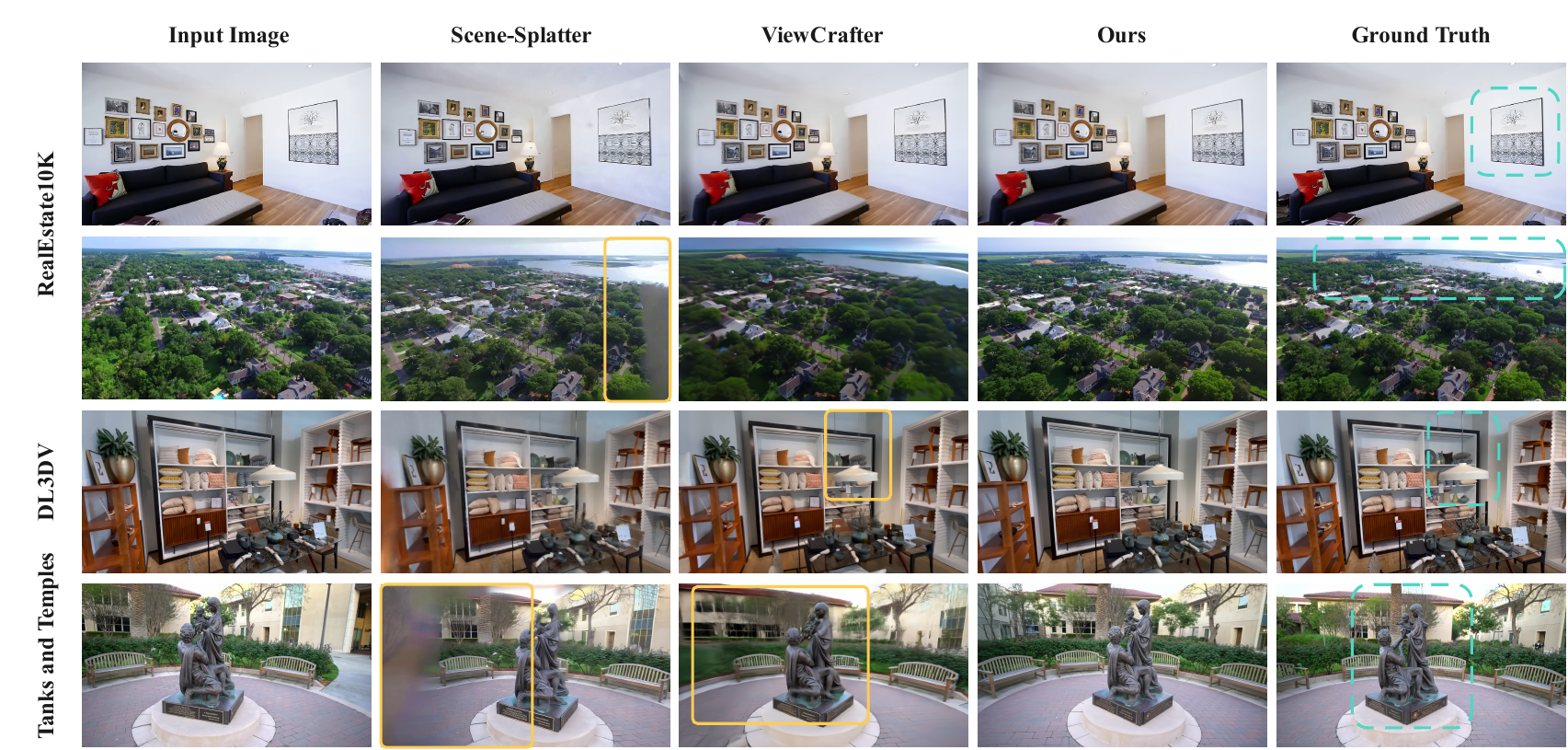} 
    \caption{Qualitative comparison with prior methods in 3D scene reconstruction, where blue bounding boxes show visible regions derived from input image and yellow bounding boxes highlight low-quality regions.}
    \label{fig:3dgs}
\end{figure*}

\subsection{Implementation Details}
\label{section:implementation}
We build our model upon the transformer-based video diffusion backbone Wan2.1~\cite{wan2025wan}, 
with LoRA modules injected into cross-attention layers to efficiently modulate spatial-temporal features. 
The model is fine-tuned for 30{,}000 iterations using the Adam optimizer with a learning rate of $1\times10^{-4}$, 
weight decay of $3\times10^{-2}$, and BF16 mixed precision. 
Input videos are divided into 81-frame clips, resized to $256\times256$, 
and encoded using a high-resolution VAE at $1024\times1024$. 
We construct the training set from two benchmark datasets with camera pose annotations: 
RealEstate10K~\cite{zhou2018stereo}, containing ~80K real-world indoor/outdoor videos with estimated trajectories, 
and DL3DV~\cite{ling2024dl3dv}, comprising 10K diverse indoor/outdoor videos with high-quality pose annotations. 
During inference, we employ the DPM solver~\cite{lu2022dpm} with 40 steps, a guidance scale of 6.5, 
and LoRA weights fixed at 0.7.

\begin{table*}[t]
\centering

\caption{Quantitative Comparison of Controllable Novel View Synthesis}
\label{tab:perceptual_metrics}
\setlength{\tabcolsep}{5pt} 
\begin{tabular}{l|ccc|ccc|ccc}
\toprule
\multirow{2}{*}{\textbf{Method}} & \multicolumn{3}{c|}{\textbf{RealEstate10K}} & \multicolumn{3}{c|}{\textbf{DL3DV}} & \multicolumn{3}{c}{\textbf{Tanks and Temples}} \\
& LPIPS $\downarrow$ & PSNR $\uparrow$ & SSIM $\uparrow$ & LPIPS $\downarrow$ & PSNR $\uparrow$ & SSIM $\uparrow$ & LPIPS $\downarrow$ & PSNR $\uparrow$ & SSIM $\uparrow$ \\
\midrule
MotionCtrl~\cite{wang2024motionctrl}     & 0.381 & 16.17 & 0.443 & 0.296 & 15.85 & 0.485 & 0.389 & 14.62 & 0.421 \\
Scene-Splatter~\cite{zhang2025scene}   & 0.332 & 17.91 & 0.506 & 0.259 & 16.80 & 0.519 & 0.345 & 15.20 & 0.489 \\
ViewCrafter~\cite{yu2024viewcrafter}    & 0.258 & 18.52 & 0.518 & 0.236 & 16.98 & 0.526 & 0.285 & 16.38 & 0.514 \\
Wonderland~\cite{liang2025wonderland}  & \underline{0.206} & \underline{19.71} & \underline{0.557} & \underline{0.218} & \underline{17.56} & \underline{0.543} & \underline{0.221} & \underline{16.87} & \underline{0.529} \\
\textbf{Ours}                          & \textbf{0.158} & \textbf{21.15} & \textbf{0.680} & \textbf{0.196} & \textbf{17.79} & \textbf{0.576} & \textbf{0.212} & \textbf{17.28} & \textbf{0.547} \\
\bottomrule
\end{tabular}
\end{table*}

\begin{table}[t]
\centering
\caption{Comparison of Distributional Metrics}
\label{tab:distributional_metrics}
\resizebox{\columnwidth}{!}{
\begin{tabular}{l|cc|cc|cc}
\toprule
\multirow{2}{*}{\textbf{Method}} & \multicolumn{2}{c|}{\textbf{RealEstate10K}} & \multicolumn{2}{c|}{\textbf{DL3DV}} & \multicolumn{2}{c}{\textbf{Tanks and Temples}} \\
& FID $\downarrow$ & FVD $\downarrow$ & FID $\downarrow$ & FVD $\downarrow$ & FID $\downarrow$ & FVD $\downarrow$ \\
\midrule
MotionCtrl~\cite{wang2024motionctrl}     & 24.12 & 255.18 & 28.43 & 292.62 & 34.65 & 327.49  \\
Scene-Splatter~\cite{zhang2025scene}   & 22.09 & 223.71 & 25.70 & 242.97 & 26.37 & 274.25 \\
ViewCrafter~\cite{yu2024viewcrafter}  & 21.28 & 208.57 & 23.46 & 236.45 & 24.48 & 256.13 \\
Wonderland~\cite{liang2025wonderland}  & \underline{16.16} & \underline{153.48}  & \underline{17.74} & \underline{169.34} & \underline{19.46} & \underline{189.32} \\
\textbf{Ours}                            & \textbf{15.88} & \textbf{143.85} & \textbf{16.86} & \textbf{158.61} & \textbf{19.13} & \textbf{181.49} \\
\bottomrule
\end{tabular}
}
\end{table}

\begin{table}[t]
\centering
\caption{Comparison of Camera Pose Errors}
\label{tab:pose_errors}
\resizebox{\columnwidth}{!}{
\begin{tabular}{l|cc|cc|cc}
\toprule
\multirow{2}{*}{\textbf{Method}} & \multicolumn{2}{c|}{\textbf{RealEstate10K}} & \multicolumn{2}{c|}{\textbf{DL3DV}} & \multicolumn{2}{c}{\textbf{Tanks and Temples}} \\
& $R_{err}$ $\downarrow$ & $T_{err}$ $\downarrow$ & $R_{err}$ $\downarrow$ & $T_{err}$ $\downarrow$ & $R_{err}$ $\downarrow$ & $T_{err}$ $\downarrow$ \\
\midrule
MotionCtrl~\cite{wang2024motionctrl}     & 0.226 & 0.664 & 0.343 & 0.862 & 0.576 & 1.207  \\
Scene-Splatter~\cite{zhang2025scene}   & 0.096 & 0.280 & 0.125 & 0.347 & 0.241 & 0.426 \\
ViewCrafter~\cite{yu2024viewcrafter}  & 0.073 & 0.194 & 0.104 & 0.216 & 0.144 & 0.337 \\
Wonderland~\cite{liang2025wonderland}  & \underline{0.046} & \underline{0.093} & \underline{0.061} & \underline{0.130} & \underline{0.094} & \underline{0.172} \\
\textbf{Ours}                            & \textbf{0.021} & \textbf{0.083} & \textbf{0.047} & \textbf{0.113} & \textbf{0.082} & \textbf{0.148} \\
\bottomrule
\end{tabular}
}
\end{table}
\subsection{Controllable Novel View Synthesis}
\label{subsectionB}
We evaluate controllable novel view synthesis in Sec. \ref{NVS} by comparing both visual generation quality and camera guidance accuracy against several baselines: 
Wonderland~\cite{liang2025wonderland}, Scene-Splatter~\cite{zhang2025scene}, ViewCrafter~\cite{yu2024viewcrafter} and MotionCtrl~\cite{wang2024motionctrl}.


\subsubsection{Comparison of Benchmark Datasets and Metrics}
We evaluate our model on three datasets: 300 test videos from RealEstate10K~\cite{zhou2018stereo}, 300 clips from DL3DV~\cite{ling2024dl3dv}, 
and 100 clips from all 14 scenes of Tanks-and-Temples~\cite{knapitsch2017tanks} for out-of-domain evaluation. 
Evaluation metrics include: (1) {Visual Similarity} measured by PSNR, SSIM, and LPIPS against ground-truth views, 
where only the first 14 frames are considered following Wonderland~\cite{liang2025wonderland} to avoid long-horizon drift (note: quantitative metrics for Wonderland are reported from the original paper as the code is not publicly available); 
(2) {Visual Quality and Temporal Coherence} assessed by FID and FVD; 
and (3) {Camera-Guidance Precision} measured by rotation error ($R_{err}$) and translation error ($T_{err}$). For the last metric, 
camera poses are recovered from generated videos using Colmap~\cite{schonberger2016structure}, 
aligned to the first frame, normalized to a unified scale, and averaged across frames under the same pose conditions.

\subsubsection{Qualitative Comparison}  
Fig.~\ref{fig:qualitative} presents qualitative results on the evaluation datasets, where the bottom-right of each input image shows the corresponding input camera frustum trajectory.  
MotionCtrl~\cite{wang2024motionctrl} generates the lowest-resolution results and exhibits the weakest trajectory alignment due to coarse camera embeddings.  
Scene-Splatter~\cite{zhang2025scene} suffers from poor geometric consistency in novel view synthesis, as it relies on a low-performance feedforward model as input condition.  
ViewCrafter~\cite{yu2024viewcrafter} produces frame-wise artifacts caused by incomplete point clouds with irregular missing regions.  
In contrast, our method achieves superior fidelity and more accurate camera control.

\begin{table*}[t]
\centering
\caption{Quantitative Comparison of 3D Scene Reconstruction}
\label{tab:2}
\setlength{\tabcolsep}{5pt}
\begin{tabular}{l|ccc|ccc|ccc}
\toprule
\multirow{2}{*}{\textbf{{Method}}} & \multicolumn{3}{c|}{\textbf{RealEstate10K}} & \multicolumn{3}{c|}{\textbf{DL3DV}} & \multicolumn{3}{c}{\textbf{Tanks-and-Temples}} \\
 & LPIPS $\downarrow$ & PSNR $\uparrow$ & SSIM $\uparrow$ & LPIPS $\downarrow$ & PSNR $\uparrow$ & SSIM $\uparrow$ & LPIPS $\downarrow$ & PSNR $\uparrow$ & SSIM $\uparrow$ \\
\midrule
Scene-Splatter~\cite{zhang2025scene} & 0.370  &  16.41 &  0.482 & 0.386  & 15.51  & 0.503  &  0.392 &  15.08 & 0.479  \\
ViewCrafter~\cite{yu2024viewcrafter}  & 0.338  & 16.88  &  0.523 & 0.364  & 15.75  & 0.529  & 0.372  & 15.26  & 0.491  \\
Wonderland~\cite{liang2025wonderland}   & \underline{0.292}  &  \underline{17.15} & \underline{0.550}  & \underline{0.325}  & \underline{16.64}  &  \underline{0.574} & \underline{0.344}  & \underline{15.90}  &  \underline{0.510} \\
\textbf{Ours}                  & \textbf{0.179} & \textbf{19.08} & \textbf{0.662} & \textbf{0.291} & \textbf{17.27} & \textbf{0.595} & \textbf{0.308} & \textbf{16.52} & \textbf{0.542} \\
\bottomrule
\end{tabular}
\end{table*}

\subsubsection{Quantitative Comparisons}
Quantitative results are reported in Table~\ref{tab:perceptual_metrics}. 
Our method consistently outperforms baselines across all metrics. Lower FID and FVD values indicate closer alignment with the ground-truth distribution. 
Smaller LPIPS and higher PSNR/SSIM confirm superior visual similarity. Furthermore, our model achieves more precise camera control, as evidenced by lower ${R}_{err}$ and ${T}_{err}$ values.

\subsection{3D Scene Reconstruction}
\label{subsection:3d_reconstruction}
We evaluate our method against several baseline approaches, including Wonderland~\cite{liang2025wonderland},
ViewCrafter~\cite{yu2024viewcrafter}, Scene-Splatter~\cite{zhang2025scene}, on real-world datasets for 3D scene generation. 
These baselines all support 3D scene generation conditioned on a single input image and camera trajectory.

\subsubsection{Comparison of Benchmark Datasets and Metrics} To evaluate 3D scene generation on benchmark datasets, we sampled 100, 100, 
and 50 images along with camera trajectories from the RealEstate10K~\cite{zhou2018stereo}, DL3DV~\cite{ling2024dl3dv}, and Tanks \& Temples~\cite{knapitsch2017tanks} test sets, respectively, 
using the sampling strategy described in Sec.~\ref{subsectionB}. For quantitative evaluation, we measured LPIPS, SSIM, 
and PSNR by comparing the renderings against ground-truth frames from the source datasets. Evaluating in this under-constrained setting is challenging, 
since multiple 3D scenes can be regarded as consistent generations for a given view~\cite{gao2024cat3d}. Therefore, 
following Sec.~\ref{subsectionB}, we used 14 sampled frames subsequent to the conditional image for metric calculation.

\subsubsection{Qualitative Comparison} The qualitative comparison in Fig. ~\ref{fig:3dgs} demonstrates the superior 3D generation capabilities of our model. 
Scene-Splatter~\cite{zhang2025scene} produces blurry renderings lacking fine details, 
while ViewCrafter~\cite{yu2024viewcrafter} improves fidelity in visible regions but struggles in handling occluded areas. 
In contrast, our model preserves intricate details and accurately reconstructs both visible and occluded regions. By leveraging priors from the video diffusion backbone, 
our approach further generates high-fidelity and visually coherent novel views, even for unseen perspectives.

\subsubsection{Quantitative Results} As shown in Tab.~\ref{tab:2}, our method significantly outperforms all baselines across multiple datasets. 
These results affirm that our model is capable of producing high-fidelity, geometrically consistent 3D scenes from single-view inputs.

\subsection{Iterative View Synthesis with Collision-Aware Camera Trajectory Planning}
\label{subsection:collision_planning}
\begin{table}[t]
\centering
\caption{Comparison of reconstruction quality and efficiency between Ours and ViewCrafter.}
\label{tab:coverage_noise}
\renewcommand{\arraystretch}{1.1}
\resizebox{\linewidth}{!}{%
\begin{tabular}{c|c c c|c}
\toprule
\textbf{Method} & Coverage (\%) $\uparrow$ & Noise Ratio $\downarrow$ & F-score@2cm $\uparrow$ & Runtime (min) $\downarrow$ \\
\midrule
ViewCrafter~\cite{yu2024viewcrafter} & 66.20 & 0.240 & 0.421 & \textbf{4.75} \\
\textbf{Ours} & \textbf{77.67} & \textbf{0.078} & \textbf{0.593} & 4.79 \\
\bottomrule
\end{tabular}%
}
\end{table}

\begin{figure}[htbp]
    \centering
    \includegraphics[width=0.48\textwidth]{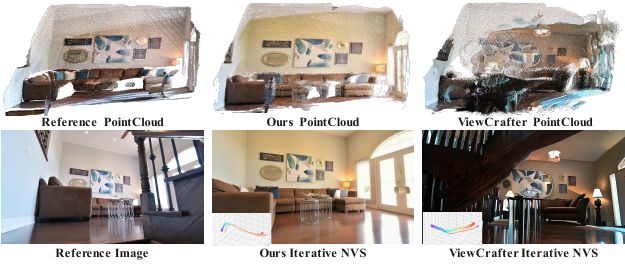}
    \caption{Comparison of reconstruction quality between Ours and ViewCrafter.}
    \label{fig:Ablation camera}
\end{figure}

We evaluated the effect of collision-aware camera trajectory planning under identical conditions as ViewCrafter~\cite{yu2024viewcrafter}, 
using the same initial point cloud, reference images, a quarter-sphere search space, and parameters $N=3$, $K=3$. 
The resulting camera trajectories from our iterative synthesis 
are visualized in the lower-left corner of each view in Fig.~\ref{fig:Ablation camera}.

ViewCrafter~\cite{yu2024viewcrafter} selects viewpoints by utility and smooth interpolation, 
often causing intersections with scene geometry and leading to fragmented point clouds. In contrast, 
our approach employs the collision detector $\mathcal{G}(\cdot)$ during sampling and interpolation, 
resolving conflicts through collision-aware optimization and yielding geometrically valid trajectories.

We evaluate four metrics using a $2\,\text{cm}$ threshold: Coverage, defined as the percentage of ground-truth points matched; 
Noise Ratio, the fraction of unmatched predictions; F-score@2cm, the harmonic mean of precision and recall; and Runtime. 
Table~\ref{tab:coverage_noise} shows that collision-aware planning substantially improves completeness and geometric accuracy while maintaining similar runtime.

\subsection{Ablation on 3D Scene Reconstruction}

\label{subsection:ablation}
\begin{figure}[htbp] 
    \centering
    \includegraphics[width=0.48\textwidth]{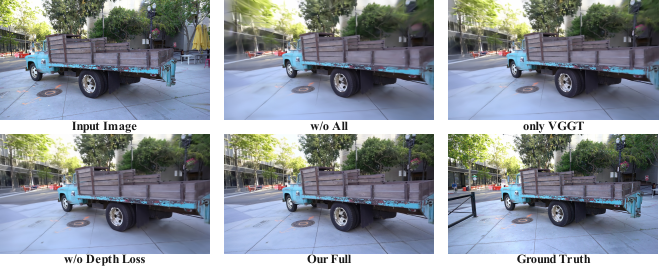} 
    \caption{Ablation study of 3D scene reconstruction. }
    \label{fig:Ablation 3D scene  reconstruction}
\end{figure}

We conduct ablation experiments on single-view 3D scene generation to evaluate the contribution of each component in our method. As shown in Fig.~\ref{fig:Ablation 3D scene reconstruction} and summarized in Table~\ref{tab:ablation}, a checkmark ($\checkmark$) indicates that the component is enabled, while a cross ($\times$) indicates its removal. Three variants are evaluated:
\textit{w/o Depth-aligned Supervision:} This variant removes the depth loss applied to calibrated monocular depths.
\textit{w/o Structural Regularization:} This variant discards the progressive Gaussian dropping mechanism.
\textit{w/o Refinement:} This variant excludes the image refinement module.
The results clearly demonstrate that the removal of any individual component leads to performance degradation, underscoring the importance of each component in ensuring high-quality 3D scene reconstruction.

\begin{table}[t]
\centering
\caption{Ablation study results of 3D Scene Reconstruction.}
\label{tab:ablation}
\resizebox{\linewidth}{!}{
\begin{tabular}{ccc|ccc}
\toprule
\makecell[c]{ Depth-aligned\\[-2pt] Supervision}  
& \makecell[c]{ Structural\\[-2pt] Regularization} 
& \makecell[c]{ Refinement} 
& LPIPS $\downarrow$ & PSNR $\uparrow$ & SSIM $\uparrow$ \\
\midrule
$\times$ & $\times$ & $\times$ & 0.341 & 15.42 & 0.471 \\
$\checkmark$ & $\times$ & $\times$ & 0.325 & 16.12 & 0.502 \\
$\checkmark$ & $\checkmark$ & $\times$ & 0.301 & 17.14 & 0.534 \\
\textbf{$\checkmark$} & \textbf{$\checkmark$} & \textbf{$\checkmark$} 
& \textbf{0.252} & \textbf{18.45} & \textbf{0.610} \\
\bottomrule
\end{tabular}
}
\end{table}

\section{Conclusion and Future Work}

In this paper, we presented \textit{NavCrafter}, a novel framework for controllable novel-view synthesis and high-fidelity 3D scene generation from a single image.
By leveraging the rich generative priors embedded in camera-conditioned video diffusion models and employing iterative view synthesis with collision-aware camera trajectory planning, our approach effectively addresses the multi-view requirements of scalable 3D scene synthesis.
The proposed multi-stage camera control architecture enables precise pose control and consistency in novel-view synthesis, 
while the geometry-aware 3D reconstruction component combines the generative capability of video diffusion models with enhanced 3D Gaussian Splatting to produce high-fidelity and geometrically consistent reconstructions.
Extensive experiments demonstrate that NavCrafter outperforms existing methods in both video generalization and 3D reconstruction quality.
For future work, we plan to address more aggressive camera motions, enhance geometric consistency over long video sequences, and extend our approach to dynamic scenes.

\bibliographystyle{IEEEtran}
\small\bibliography{reference}

\begin{thebibliography}{10}
\providecommand{\url}[1]{#1}
\csname url@samestyle\endcsname
\providecommand{\newblock}{\relax}
\providecommand{\bibinfo}[2]{#2}
\providecommand{\BIBentrySTDinterwordspacing}{\spaceskip=0pt\relax}
\providecommand{\BIBentryALTinterwordstretchfactor}{4}
\providecommand{\BIBentryALTinterwordspacing}{\spaceskip=\fontdimen2\font plus
\BIBentryALTinterwordstretchfactor\fontdimen3\font minus
  \fontdimen4\font\relax}
\providecommand{\BIBforeignlanguage}[2]{{%
\expandafter\ifx\csname l@#1\endcsname\relax
\typeout{** WARNING: IEEEtran.bst: No hyphenation pattern has been}%
\typeout{** loaded for the language `#1'. Using the pattern for}%
\typeout{** the default language instead.}%
\else
\language=\csname l@#1\endcsname
\fi
#2}}
\providecommand{\BIBdecl}{\relax}
\BIBdecl

\bibitem{mildenhall2021nerf}
B.~Mildenhall, P.~P. Srinivasan, M.~Tancik, J.~T. Barron, R.~Ramamoorthi, and
  R.~Ng, ``Nerf: Representing scenes as neural radiance fields for view
  synthesis,'' \emph{Communications of the ACM}, vol.~65, no.~1, pp. 99--106,
  2021.

\bibitem{kerbl20233d}
B.~Kerbl, G.~Kopanas, T.~Leimk{\"u}hler, and G.~Drettakis, ``3d gaussian
  splatting for real-time radiance field rendering.'' \emph{ACM Trans. Graph.},
  vol.~42, no.~4, pp. 139--1, 2023.

\bibitem{ho2020denoising}
J.~Ho, A.~Jain, and P.~Abbeel, ``Denoising diffusion probabilistic models,''
  \emph{Advances in neural information processing systems}, vol.~33, pp.
  6840--6851, 2020.

\bibitem{song2020score}
Y.~Song, J.~Sohl-Dickstein, D.~P. Kingma, A.~Kumar, S.~Ermon, and B.~Poole,
  ``Score-based generative modeling through stochastic differential
  equations,'' \emph{arXiv preprint arXiv:2011.13456}, 2020.

\bibitem{chung2023luciddreamer}
J.~Chung, S.~Lee, H.~Nam, J.~Lee, and K.~M. Lee, ``Luciddreamer: Domain-free
  generation of 3d gaussian splatting scenes,'' \emph{arXiv preprint
  arXiv:2311.13384}, 2023.

\bibitem{yu2025wonderworld}
H.-X. Yu, H.~Duan, C.~Herrmann, W.~T. Freeman, and J.~Wu, ``Wonderworld:
  Interactive 3d scene generation from a single image,'' in \emph{Proceedings
  of the Computer Vision and Pattern Recognition Conference}, 2025, pp.
  5916--5926.

\bibitem{blattmann2023stable}
A.~Blattmann, T.~Dockhorn, S.~Kulal, D.~Mendelevitch, M.~Kilian, D.~Lorenz,
  Y.~Levi, Z.~English, V.~Voleti, A.~Letts \emph{et~al.}, ``Stable video
  diffusion: Scaling latent video diffusion models to large datasets,''
  \emph{arXiv preprint arXiv:2311.15127}, 2023.

\bibitem{wang2024motionctrl}
Z.~Wang, Z.~Yuan, X.~Wang, Y.~Li, T.~Chen, M.~Xia, P.~Luo, and Y.~Shan,
  ``Motionctrl: A unified and flexible motion controller for video
  generation,'' in \emph{ACM SIGGRAPH 2024 Conference Papers}, 2024, pp. 1--11.

\bibitem{he2025cameractrl}
H.~He, Y.~Xu, Y.~Guo, G.~Wetzstein, B.~Dai, H.~Li, and C.~Yang, ``Cameractrl:
  Enabling camera control for video diffusion models,'' in \emph{The Thirteenth
  International Conference on Learning Representations}, 2025.

\bibitem{ren2025gen3c}
X.~Ren, T.~Shen, J.~Huang, H.~Ling, Y.~Lu, M.~Nimier-David, T.~M{\"u}ller,
  A.~Keller, S.~Fidler, and J.~Gao, ``Gen3c: 3d-informed world-consistent video
  generation with precise camera control,'' in \emph{Proceedings of the
  Computer Vision and Pattern Recognition Conference}, 2025, pp. 6121--6132.

\bibitem{chen2024liftimage3d}
Y.~Chen, C.~Yang, J.~Fang, X.~Zhang, L.~Xie, W.~Shen, W.~Dai, H.~Xiong, and
  Q.~Tian, ``Liftimage3d: Lifting any single image to 3d gaussians with video
  generation priors,'' \emph{arXiv preprint arXiv:2412.09597}, 2024.

\bibitem{park2025dropgaussian}
H.~Park, G.~Ryu, and W.~Kim, ``Dropgaussian: Structural regularization for
  sparse-view gaussian splatting,'' in \emph{Proceedings of the Computer Vision
  and Pattern Recognition Conference}, 2025, pp. 21\,600--21\,609.

\bibitem{li2024dngaussian}
J.~Li, J.~Zhang, X.~Bai, J.~Zheng, X.~Ning, J.~Zhou, and L.~Gu, ``Dngaussian:
  Optimizing sparse-view 3d gaussian radiance fields with global-local depth
  normalization,'' in \emph{Proceedings of the IEEE/CVF conference on computer
  vision and pattern recognition}, 2024, pp. 20\,775--20\,785.

\bibitem{chen2024mvsplat360}
Y.~Chen, C.~Zheng, H.~Xu, B.~Zhuang, A.~Vedaldi, T.-J. Cham, and J.~Cai,
  ``Mvsplat360: Feed-forward 360 scene synthesis from sparse views,''
  \emph{Advances in Neural Information Processing Systems}, vol.~37, pp.
  107\,064--107\,086, 2024.

\bibitem{ma2025you}
B.~Ma, H.~Gao, H.~Deng, Z.~Luo, T.~Huang, L.~Tang, and X.~Wang, ``You see it,
  you got it: Learning 3d creation on pose-free videos at scale,'' in
  \emph{Proceedings of the Computer Vision and Pattern Recognition Conference},
  2025, pp. 2016--2029.

\bibitem{gao2024cat3d}
R.~Gao, A.~Holynski, P.~Henzler, A.~Brussee, R.~Martin-Brualla, P.~Srinivasan,
  J.~T. Barron, and B.~Poole, ``Cat3d: Create anything in 3d with multi-view
  diffusion models,'' \emph{arXiv preprint arXiv:2405.10314}, 2024.

\bibitem{yu2024viewcrafter}
W.~Yu, J.~Xing, L.~Yuan, W.~Hu, X.~Li, Z.~Huang, X.~Gao, T.-T. Wong, Y.~Shan,
  and Y.~Tian, ``Viewcrafter: Taming video diffusion models for high-fidelity
  novel view synthesis,'' \emph{arXiv preprint arXiv:2409.02048}, 2024.

\bibitem{xiao20243dtrajmaster}
F.~Xiao, X.~Liu, X.~Wang, S.~Peng, M.~Xia, X.~Shi, Z.~Yuan, P.~Wan, D.~Zhang,
  and D.~Lin, ``3dtrajmaster: Mastering 3d trajectory for multi-entity motion
  in video generation,'' in \emph{The Thirteenth International Conference on
  Learning Representations}, 2024.

\bibitem{hu2024motionmaster}
T.~Hu, J.~Zhang, R.~Yi, Y.~Wang, H.~Huang, J.~Weng, Y.~Wang, and L.~Ma,
  ``Motionmaster: Training-free camera motion transfer for video generation,''
  \emph{arXiv preprint arXiv:2404.15789}, 2024.

\bibitem{hu2022lora}
E.~J. Hu, Y.~Shen, P.~Wallis, Z.~Allen-Zhu, Y.~Li, S.~Wang, L.~Wang, W.~Chen
  \emph{et~al.}, ``Lora: Low-rank adaptation of large language models.''
  \emph{ICLR}, vol.~1, no.~2, p.~3, 2022.

\bibitem{sun2024dimensionx}
W.~Sun, S.~Chen, F.~Liu, Z.~Chen, Y.~Duan, J.~Zhang, and Y.~Wang, ``Dimensionx:
  Create any 3d and 4d scenes from a single image with controllable video
  diffusion,'' in \emph{International Conference on Computer Vision (ICCV)},
  2025.

\bibitem{liang2025wonderland}
H.~Liang, J.~Cao, V.~Goel, G.~Qian, S.~Korolev, D.~Terzopoulos, K.~N.
  Plataniotis, S.~Tulyakov, and J.~Ren, ``Wonderland: Navigating 3d scenes from
  a single image,'' in \emph{Proceedings of the Computer Vision and Pattern
  Recognition Conference}, 2025, pp. 798--810.

\bibitem{zhai2025stargen}
S.~Zhai, Z.~Ye, J.~Liu, W.~Xie, J.~Hu, Z.~Peng, H.~Xue, D.~Chen, X.~Wang,
  L.~Yang \emph{et~al.}, ``Stargen: A spatiotemporal autoregression framework
  with video diffusion model for scalable and controllable scene generation,''
  in \emph{Proceedings of the Computer Vision and Pattern Recognition
  Conference}, 2025, pp. 26\,822--26\,833.

\bibitem{yu2024wonderjourney}
H.-X. Yu, H.~Duan, J.~Hur, K.~Sargent, M.~Rubinstein, W.~T. Freeman, F.~Cole,
  D.~Sun, N.~Snavely, J.~Wu \emph{et~al.}, ``Wonderjourney: Going from anywhere
  to everywhere,'' in \emph{Proceedings of the IEEE/CVF Conference on Computer
  Vision and Pattern Recognition}, 2024, pp. 6658--6667.

\bibitem{wan2025wan}
T.~Wan, A.~Wang, B.~Ai, B.~Wen, C.~Mao, C.-W. Xie, D.~Chen, F.~Yu, H.~Zhao,
  J.~Yang \emph{et~al.}, ``Wan: Open and advanced large-scale video generative
  models,'' \emph{arXiv preprint arXiv:2503.20314}, 2025.

\bibitem{wang2025vggt}
J.~Wang, M.~Chen, N.~Karaev, A.~Vedaldi, C.~Rupprecht, and D.~Novotny, ``Vggt:
  Visual geometry grounded transformer,'' in \emph{Proceedings of the Computer
  Vision and Pattern Recognition Conference}, 2025, pp. 5294--5306.

\bibitem{wang2025moge}
R.~Wang, S.~Xu, Y.~Dong, Y.~Deng, J.~Xiang, Z.~Lv, G.~Sun, X.~Tong, and
  J.~Yang, ``Moge-2: Accurate monocular geometry with metric scale and sharp
  details,'' \emph{arXiv preprint arXiv:2507.02546}, 2025.

\bibitem{wu2025difix3d+}
J.~Z. Wu, Y.~Zhang, H.~Turki, X.~Ren, J.~Gao, M.~Z. Shou, S.~Fidler, Z.~Gojcic,
  and H.~Ling, ``Difix3d+: Improving 3d reconstructions with single-step
  diffusion models,'' in \emph{Proceedings of the Computer Vision and Pattern
  Recognition Conference}, 2025, pp. 26\,024--26\,035.

\bibitem{zhou2018stereo}
T.~Zhou, R.~Tucker, J.~Flynn, G.~Fyffe, and N.~Snavely, ``Stereo magnification:
  Learning view synthesis using multiplane images,'' \emph{arXiv preprint
  arXiv:1805.09817}, 2018.

\bibitem{ling2024dl3dv}
L.~Ling, Y.~Sheng, Z.~Tu, W.~Zhao, C.~Xin, K.~Wan, L.~Yu, Q.~Guo, Z.~Yu, Y.~Lu
  \emph{et~al.}, ``Dl3dv-10k: A large-scale scene dataset for deep
  learning-based 3d vision,'' in \emph{Proceedings of the IEEE/CVF Conference
  on Computer Vision and Pattern Recognition}, 2024, pp. 22\,160--22\,169.

\bibitem{lu2022dpm}
C.~Lu, Y.~Zhou, F.~Bao, J.~Chen, C.~Li, and J.~Zhu, ``Dpm-solver: A fast ode
  solver for diffusion probabilistic model sampling in around 10 steps,''
  \emph{Advances in neural information processing systems}, vol.~35, pp.
  5775--5787, 2022.

\bibitem{zhang2025scene}
S.~Zhang, J.~Li, X.~Fei, H.~Liu, and Y.~Duan, ``Scene splatter: Momentum 3d
  scene generation from single image with video diffusion model,'' in
  \emph{Proceedings of the Computer Vision and Pattern Recognition Conference},
  2025, pp. 6089--6098.

\bibitem{knapitsch2017tanks}
A.~Knapitsch, J.~Park, Q.-Y. Zhou, and V.~Koltun, ``Tanks and temples:
  Benchmarking large-scale scene reconstruction,'' \emph{ACM Transactions on
  Graphics (ToG)}, vol.~36, no.~4, pp. 1--13, 2017.

\bibitem{schonberger2016structure}
J.~L. Schonberger and J.-M. Frahm, ``Structure-from-motion revisited,'' in
  \emph{Proceedings of the IEEE conference on computer vision and pattern
  recognition}, 2016, pp. 4104--4113.

\end{thebibliography}

\end{document}